\documentclass{svproc}
\usepackage{url}

\usepackage{graphicx}
\usepackage{tabularx}
\usepackage{amsmath}

\begin{document}
\mainmatter              
\title{Comparison and Evaluation of Different Simulation Environments for Rigid Body Systems}
\titlerunning{Comparison and Evaluation of Different Simulation Environments...}  
%
\author{Longxiang Shao \and Ulrich Dahmen \and Jürgen Roßmann}
\authorrunning{Longxiang Shao et al.} 
\institute{Institute for Man-Machine Interaction, RWTH Aachen University, \\ Im Süsterfeld 9, 52072 Aachen, Germany,\\
\email{shao@mmi.rwth-aachen.de}
}

\maketitle              

\begin{abstract}
Rigid body dynamics simulators are important tools for the design, analysis and optimization of mechanical systems in a variety of technical and scientific applications. This study examines four different simulation environments (Adams, Simscape, OpenModelica, and VEROSIM), focusing in particular on the comparison of the modeling methods, the numerical solvers, and the treatment of numerical problems that arise especially in closed-loop kinematics (esp. redundant boundary conditions and static equilibrium problem). A novel and complex crane boom of a real forestry machine serves as a practical benchmark application example. The direct comparison of the different solution approaches in the examined simulation tools supports the user in selecting the most suitable tool for his application.
%
\end{abstract}
\section{Introduction}
Rigid body dynamics simulators are indispensable tools for analyzing and optimizing mechanical systems in various engineering and scientific domains, including robotics, industrial machinery, and aerospace. Numerical simulations allow engineers and researchers to predict system behavior, validate designs, and optimize performance before physical implementation, thereby reducing development time and costs. As engineering challenges become increasingly complex, the demand for accurate, efficient, and user-friendly simulation environments has significantly increased.

Numerous simulation environments have been developed to address this need, each employing different approaches to modeling and solving rigid body dynamics. Among these, Adams, Simscape, and OpenModelica are widely recognized for their capabilities and applications in academia and industry \cite{ryan,poz:ach:val}. VEROSIM \cite{rossmann2012}, developed at RWTH Aachen University, is less commonly known but offers some interesting features. The mentioned tools exhibit substantial differences in their modeling frameworks, numerical solvers, user interfaces, and ability to handle complex systems. Consequently, selecting the most suitable simulation environment for a specific application requires a thorough understanding of their strengths, limitations, and applicability to specific problems that arise in real-world scenarios.

In previous work, the applicability of different physics simulators to various research domains and simulation scenarios has been extensively discussed \cite{col:cha:van}. Widely used engines such as Bullet, Havok, MuJoCo, ODE, and PhysX have been compared for their performance in handling numerical computation challenges \cite{ere:tas:tod}. Additionally, ODE, Bullet, Vortex, and MuJoCo have been further analyzed under both simple and complex industrial conditions, providing insight into their capabilities and limitations \cite{yoo:son:lee}. Although prior studies have evaluated individual simulation tools and numerical solvers, comprehensive comparisons that systematically benchmark multiple environments under consistent criteria are still limited. Moreover, existing evaluations often do not address the advanced modeling capabilities required for complex engineering applications.

This study bridges these gaps by comparing and evaluating four simulation environments: Adams, Simscape, OpenModelica, and VEROSIM. The analysis focuses on their modeling and solving approaches for rigid body dynamics, as well as their performance in handling numerical challenges in closed-loop systems. The findings provide practical insights into their strengths and limitations, aiding users in selecting the most suitable tool. 

The remainder of this paper is organized as follows: Section 2 summarizes the mathematical foundations of the rigid body dynamics. Section 3 offers an overview of the four selected simulation environments, highlighting their key features and functionalities. Section 4 presents a comparative analysis of these tools, focusing on their respective modeling frameworks and numerical solvers. Finally, Section 5 concludes the paper with a summary of the results.

\section{Mathematical Foundations in Rigid Body Dynamics Simulation}
The state of a rigid body system can be described using two representation methods: the generalized coordinate system and the maximal coordinate system \cite{verosim}. 
The generalized coordinate method employs a set of independent variables to describe each degree of freedom in the system, encompassing both translational and rotational motions. By reducing redundant degrees of freedom, the size of the system matrix is significantly reduced. This method is particularly well-suited for systems with serial kinematic chains, such as those found in industrial robotics. 
However, challenges arise when the system involves redundant constraints, such as closed kinematic loops or multiple contact points, as the generalized coordinates lose their independence under these conditions \cite{feat}. 



The maximal coordinate method, on the other hand, explicitly describes all six degrees of freedom for each rigid body in the system. To constrain the motion of the system, additional constraint conditions are imposed between the rigid bodies. The system's equation of motion is expressed via equation \eqref{eq:maximal}

\begin{equation}
  \mathbf{M} \cdot \ddot{\vec{x}} = \vec{f}_{ext} + \vec{f}_{c} \label{eq:maximal}
\end{equation}
where $\mathbf{M}$ is the system mass matrix, $\vec{f}_{ext}$ denotes external forces, and $\vec{f}_{c}$ represents constraint forces.

This method can handle systems with arbitrary redundant constraints and is inherently suited for setups with variable bases or frequent changes in contact configurations.
However, its main drawback is the larger number of coordinates required, which can lead to numerical drift and increased computational time for constraint force calculations. Stabilization techniques are often required to address this issue. Despite this challenge, Baraff \cite{bara} demonstrated that constraint forces can still be computed in linear time for simple kinematic chains.


In multibody dynamics simulation, both holonomic constraints (e.g., those imposed by common joint types) and nonholonomic constraints (e.g., contact and friction) must be incorporated into the system, leading to differential-algebraic equations (DAEs). The maximal coordinate method commonly uses linear complementarity problems (LCPs) to model constraints, with the Open Dynamics Engine \cite{ode} being one of the most well-known dynamics engines. In contrast, the generalized coordinate method reformulates LCP as a convex quadratic programming (QP) problem, as exemplified by MuJoCo \cite{tod:ere:tas,mujoco}. Both LCPs and QPs require numerical solvers to compute the constraint forces. The Projected Gauss-Seidel (PGS) method is among the most widely used solvers due to its simplicity and efficiency. To mitigate numerical drift, Baumgarte stabilization is often employed to minimize errors during simulation. Once constraint forces are determined, forward dynamics is used to compute system accelerations. Subsequently, velocities and positions are updated using semi-implicit or implicit integration methods, which are commonly employed to enhance numerical stability and improve accuracy in handling constrained systems~\cite{mujoco}.


\section{Selected Simulation Environments}

To leverage the strengths of the maximal coordinate method in addressing systems with redundant constraints, this study focuses on four simulation environments that adopt this approach: Adams, Simscape, OpenModelica, and VEROSIM. This section provides a comprehensive overview of these four simulation environments from the perspective of rigid body dynamics simulation.

Adams (Automatic Dynamic Analysis of Mechanical Systems), developed by MSC Software, is a widely used multibody dynamics simulation software tailored for professional engineering applications. It provides highly customizable tools for modeling and analysis, supporting complex geometric designs, constraint definitions, and advanced analysis setups. Its post-processing capabilities allow users to generate detailed curves, animations, and reports. Furthermore, Adams seamlessly integrates with CAD software, facilitating efficient model import and geometric optimization.
Adams includes multiple solvers, such as GSTIFF \cite{GSTIFF}, which enable error control at both displacement and velocity levels. GSTIFF is highly effective for stiff and constrained multibody systems but velocities and especially accelerations can have 
errors \cite{adams}. Adams includes IMPACT and POISSON for calculating the contact normal forces and supports Coulomb and stiction friction models \cite{adams}. Specifically, the contact normal force calculation models provided by Adams utilize penalty regularization methods, which are simple and intuitive to implement without introducing additional equations or variables; however, their primary drawback is the requirement for users to manually set suitable penalty parameters. Additionally, the absence of real-time collision detection capabilities may limit its applicability in scenarios requiring dynamic interactions.

Simscape, built on MATLAB/Simulink, employs a drag-and-drop graphical interface for connecting components, enabling rapid development of system models. It offers pre-defined modules across various physical domains, such as mechanical, electrical, and hydraulic, simplifying multi-domain system modeling. Users can customize parameters and logic directly via MATLAB scripts or Simulink's model interface. Post-processing tools are provided for analyzing simulation results, which can also be further explored and visualized within the MATLAB environment.
Simscape transforms DAEs into ordinary differential equations (ODEs) for numerical solving \cite{simscape}. It directly utilizes MATLAB solvers, such as ODE45, offering both fixed-step and variable-step integration methods to meet diverse simulation requirements. For contact force computation, Simscape utilizes the penalty method and incorporates a smooth stick-slip friction model. However, it lacks built-in collision detection algorithms, requiring users to predefine contact pairs for accurate modeling.

OpenModelica, an open-source simulation tool, features a simple interface and supports extensive community-driven extensions and plugins. It offers a basic graphical user interface (OMEdit) that supports both graphical modeling and text-based Modelica programming. While it includes essential post-processing tools for visualizing result curves, its graphical capabilities are more limited compared to other simulation environments.
The default solver in OpenModelica, DASSL, is an implicit, high-order, multi-step solver based on the Backward Differentiation Formula \cite{dassl}. With step-size control, DASSL ensures robust stability for a wide range of models, making it suitable for various dynamic systems. However, OpenModelica does not provide built-in support for contact or friction calculations, nor does it include native collision detection capabilities.

VEROSIM (Virtual Environments and Robotics Simulation System) effectively handles redundant constraints and closed-loop mechanisms, maintaining numerical stability even in systems with high stiffness or singular configurations. It features real-time interactive simulation capabilities, allowing users to dynamically adjust parameters and observe immediate feedback through high-resolution 3D visualization. Its modular architecture supports domain-specific customization and the import of CAD geometries, streamlining the transition from design to simulation \cite{verosim}. These features make VEROSIM particularly suited for applications such as rapid prototyping, operator training, and system optimization.
VEROSIM employs impulse-based simulation modeling and utilizes the PGS method for solving dynamic equations. It supports various collision detection algorithms and offers a range of contact and friction models to address diverse simulation requirements. Additionally, the implementation of the constraint force mixing (CFM) method enables the use of soft constraints, enhancing numerical stability, but at the cost of strictly enforcing constraints, which can lead to minor penetrations, slight contact gaps, or small deviations in joint alignment \cite{ode}.
\section{Comparison of Modeling and Simulation Approaches}
\subsection{Experimental Setup}
The dynamics of simple rigid-bodies are relatively straightforward to describe, and the introduction of articulated joints is manageable for open-loop systems. However, closed-loop kinematic systems significantly increase modeling and computational complexity. For this reason, a closed-loop system was chosen as the benchmark for the comparative study conducted in this paper.

The definition and simulation of closed-loop systems often involve challenges that complicate their numerical representation and solution. Different simulation environments adopt varied approaches to address these challenges. This paper focuses on three key phenomena arising in closed-loop systems: a) redundant boundary conditions, b) the propagation and computation of constraint forces, and c) the static equilibrium problem.

\begin{figure}[h] 
    \centering
    \includegraphics[width=0.99\textwidth]{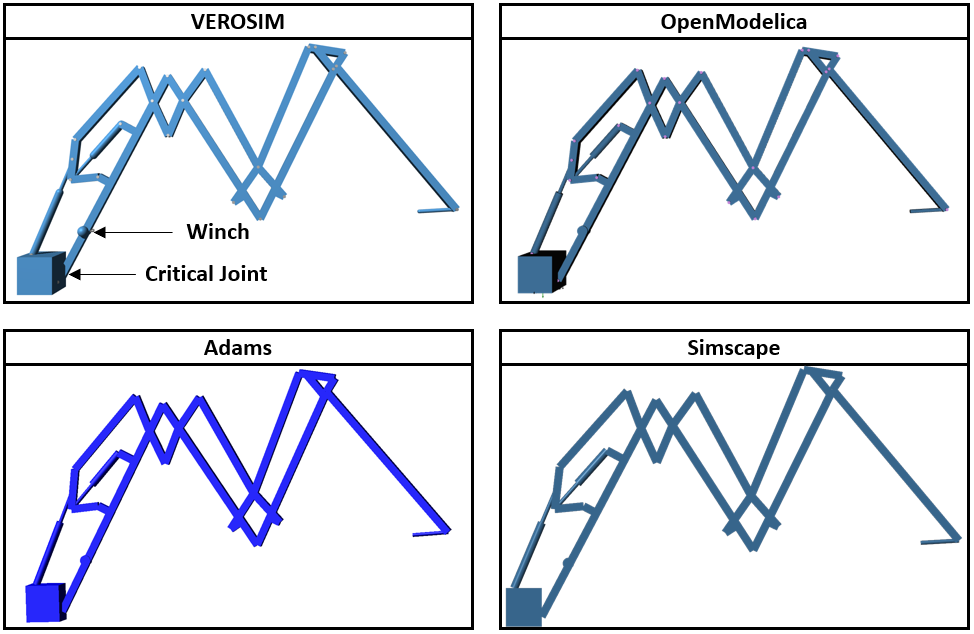}
    \caption{Visualization of the MCrane model in VEROSIM, OpenModelica, Adams, and Simscape.}
    \label{fig:MCrane_all} 
\end{figure}

As an academic example, a four-bar linkage system was modeled in all four simulation environments to analyze their handling of redundant boundary conditions. To explore the propagation and computation of constraint forces and the static equilibrium problem, a more complex real-world system was selected: A novel crane boom design, referred to as MCrane, which was developed for a forestry machine as part of a research project (see Figure \ref{fig:MCrane_all}). Characterized by multiple kinematic loops and high topological complexity, the MCrane model serves as a robust benchmark for assessing the capabilities and limitations of the selected simulation environments. Specifically, the MCrane consists of 21 discrete components (one of which remains in static equilibrium), 27 joints (25 revolute and 2 prismatic joints), and two actuators. Together, they form 9 closed and coupled kinematic loops. Figure \ref{fig:MCrane_all} shows the 3D visualizations of the respective MCrane model in VEROSIM, OpenModelica, Adams, and Simscape. All four models include the same rigid body geometries, masses, poses, and joint configurations, as well as identical control signals applied to the actuators. 
%
%
%
\subsection{Dealing with Redundant Boundary Conditions}
The processes for defining and modeling rigid bodies and open-loop mechanisms across the four simulation environments are largely similar, despite minor differences in specific methods and workflows. Typically, modeling begins with the definition of a global coordinate system as a reference for subsequent components. The rigid bodies are then defined by specifying their physical properties, including mass, inertia tensor, and geometry, as well as their initial poses. Joint modeling follows similar principles across environments, requiring the definition of joint types and the poses of the joints relative to the connected rigid bodies. Accurate alignment of these poses is critical to ensure kinematic consistency and motion continuity.

Closed-loop systems introduce redundant constraints into the multibody dynamics model, reducing the effective degrees of freedom and potentially leading to singularities or ill-conditioned system equations \cite{adams}. This redundancy can result in numerical difficulties and solver failures during simulation.

\begin{figure}[h] 
    \centering
    \includegraphics[width=0.99\textwidth]{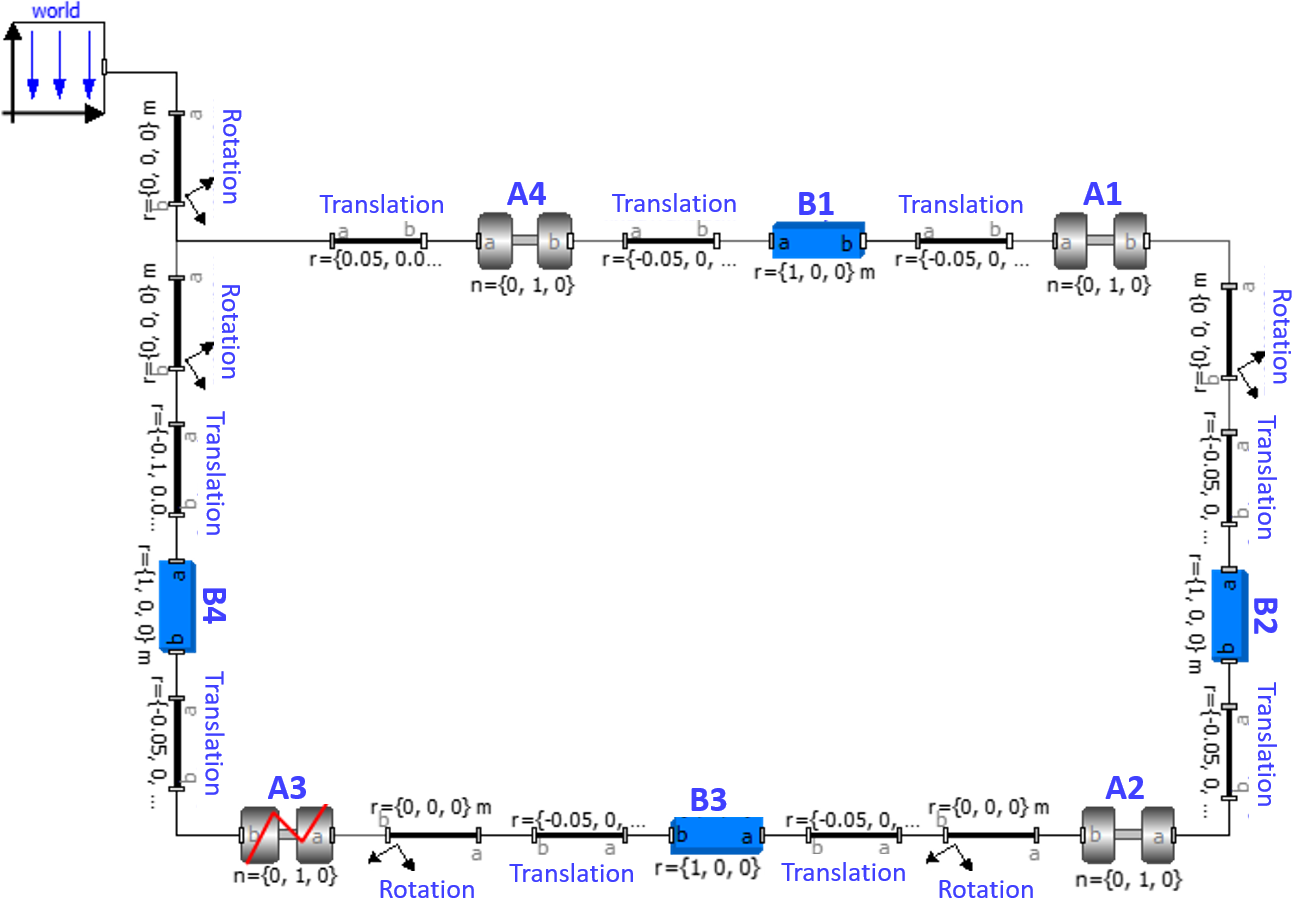}
    \caption{Modeling structure of a four-bar linkage in OpenModelica.}
    \label{fig:closed_loop_modelica} 
\end{figure}

To address this issue, different simulation tools adopt different approaches. Adams Solver and Simscape automatically detect redundant constraints and determine which constraints to remove from the equation set. However, the specific constraints removed are not disclosed to the user, which can complicate debugging and validation \cite{adams,simscape}. In contrast, OpenModelica cannot automatically detect closed-loops or redundant constraints. Users must manually identify closed-loops during the modeling process and eliminate redundant degrees of freedom by applying specialized components, such as the special revolute joint A3 illustrated in Figure \ref{fig:closed_loop_modelica}.

VEROSIM instead takes a different approach by directly solving systems with redundant constraints. It introduces relaxation variables via the CFM technique, allowing small deviations in constraint conditions \cite{verosim}. This approach mitigates numerical issues associated with redundant constraints, such as solver instability or non-convergence. By permitting slight constraint violations, VEROSIM can effectively handle high-precision dynamic systems without requiring explicit user intervention to manage closed-loops. However, to balance the required accuracy and computational stability, careful selection and tuning of the CFM parameters are necessary to avoid excessive sacrifices in physical fidelity.

In multibody dynamics systems, when the defined geometric constraints cannot be simultaneously satisfied, contradictions or errors occur in the mathematical model or simulation, leading to the inability to find a feasible solution. Therefore, it is crucial during the modeling process to ensure that all geometric constraints are fully satisfied at the initial moment to prevent such issues.

When modeling closed-loops using OpenModelica and Simscape, the precision of the initial geometric positions significantly impacts the numerical solver's performance. This is primarily due to their modeling approach, where each rigid body is defined by its two endpoint frames, FrameA and FrameB. The FrameB of a rigid body is computed relative to FrameA using a vector that specifies both the direction and distance from FrameA to FrameB, resolved in FrameA. This vector must be manually calculated by the user, which increases the modeling effort and introduces potential numerical errors.

In the system, the FrameA of a joint connects to the FrameB of the preceding rigid body, meaning these two frames must align perfectly. Similarly, the FrameB of the joint connects to the FrameA of the subsequent rigid body, thereby forming a chain of rigid bodies and joints. A closed-loop system is established when the FrameB of the last rigid body aligns precisely with the FrameA of the first rigid body.
However, due to the modeling approach in OpenModelica and Simscape, the calculation of each FrameB accumulates numerical errors stemming from the manually defined vector between FrameA and FrameB. These errors propagate along the kinematic chain and accumulate around the closed-loop structure. If the accumulated error becomes significant, the FrameB of the last rigid body will not align perfectly with the FrameA of the first rigid body, leading to numerical solver failure. 

In the simple four-bar linkage test, the initial position accuracy had no significant impact across all four simulation environments, and an initial position accuracy of $10^{-3}$m was sufficient to ensure solver stability. However, testing with practical models such as the MCrane demonstrates that the initial position accuracy must be maintained to within $10^{-6}$m to ensure stable numerical solutions in OpenModelica and Simscape.

VEROSIM and Adams adopt a different approach to modeling dynamic chains. In their frameworks, joints require only a single coordinate system for definition, eliminating the need to define two end-point frames for each joint and ensuring their alignment with the connected rigid bodies, as is required in OpenModelica and Simscape. Additionally, in VEROSIM and Adams, each rigid body is defined using a single local coordinate system and its geometric information. This local coordinate system can be defined relative to the world coordinate system, removing the necessity for strict consistency between the local coordinate systems of consecutive rigid bodies in the dynamic chain.

This modeling approach significantly reduces the influence of initial position errors on numerical solvers. Specifically, in VEROSIM and Adams, the solver does not interpret initial position discrepancies caused by precision limitations as violations of constraints, which is a common issue in OpenModelica and Simscape. Consequently, the impact of initial position accuracy on solver stability is greatly minimized. Testing with the MCrane model demonstrates that an initial position accuracy of $10^{-3}$m is sufficient to ensure stable numerical solutions in VEROSIM and Adams, which is substantially more tolerant compared to OpenModelica and Simscape.

\subsection{Dealing with Constraint Forces}
%

\begin{figure}[h] 
    \centering
    \includegraphics[width=0.95\textwidth]{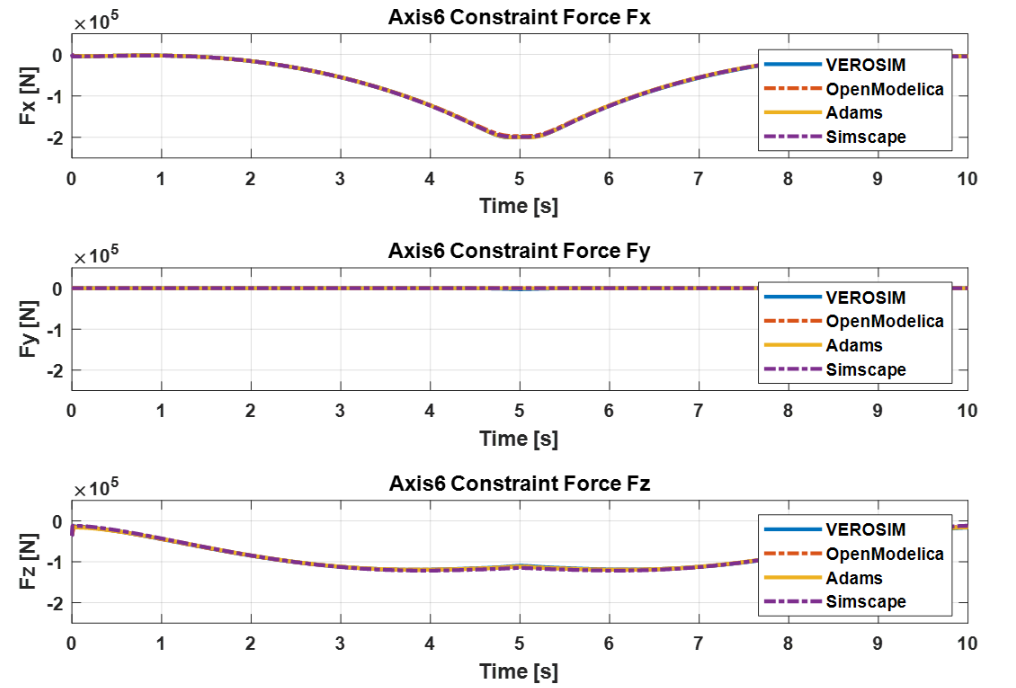}
    \caption{Comparison of constraint forces at critical joint over time across VEROSIM, OpenModelica, Adams, and Simscape}
    \label{fig:Axis6_ConstraintForce} 
\end{figure}

In multibody dynamics simulations, numerical errors propagate differently across computed quantities such as constraint forces, accelerations, velocities, and positions due to their distinct computational dependencies. At each integration step, constraint forces and accelerations are directly computed by solving coupled dynamics and constraint equations, making them highly sensitive to numerical inaccuracies, redundant constraints, and ill-conditioned matrices.
On the other hand, velocities and positions are obtained through numerical integration, and their errors accumulate progressively throughout the simulation. This error propagation mechanism underscores the critical importance of accurately computing constraint forces to minimize error accumulation and ensure numerical stability. The rotational joint that connects the main arm of the crane to the base (see Figure \ref{fig:MCrane_all}) is considered to be the most critical joint for constraint force computation. Figure \ref{fig:Axis6_ConstraintForce} demonstrates that the constraint forces at the critical joint of the MCrane model remain largely consistent across the four simulation environments.
\subsection{Dealing with Static Equilibrium Problem}
When a cylinder is connected to another rigid body via a hinge along its principal axis and the cylinder is at rest with no external forces or torques applied, its state is referred to as an equilibrium state. In this equilibrium, the absence of external influences to define a specific rotational direction renders the system's motion indeterminate, allowing the cylinder to potentially rotate either clockwise or counterclockwise. This phenomenon is known as the static equilibrium problem in dynamics, where the mathematical description of the system fails to uniquely determine its evolution \cite{equil}.

In the absence of external forces, the velocity and acceleration of the system remain zero. Since the cylinder's rotation about its principal axis does not involve any driving force, all torque terms in the governing dynamic equations vanish, resulting in a homogeneous system of equations where the right-hand side is zero. Under these conditions, the mass matrix $M(q)$ may become singular along specific directions, indicating that the determinant of the matrix is zero. This singularity reflects the fact that the system's state equations lack a unique solution, as the mathematical formulation cannot determine the cylinder's rotational direction without additional input. Consequently, the cylinder's motion is entirely dictated by initial conditions or external perturbations.

In the MCrane simulation example, a winch mechanism was designed to pull the forestry machine's front arm into a retracted position by winding and unwinding the cable. The winch was approximated as a cylinder connected to the main arm via a hinge, introducing the static equilibrium problem into the system, and is shown in Figure \ref{fig:MCrane_all}. Testing revealed that this issue could be handled by Adams, VEROSIM, and Simscape, as these tools incorporate damping or elastic elements at joints or connections to provide additional torques or forces. These stabilizing elements effectively guide the system's motion, resolving the indeterminacy. However, OpenModelica was unable to compute this scenario, as the system matrix exhibited singularity. To address this issue in OpenModelica, users must manually introduce small external perturbations or initial conditions to break the symmetry and assist the system in selecting a specific motion path.

\section{Conclusion}
In this study, we analyzed and compared the modeling approaches, numerical solvers, and performance of four simulation environments in handling systems with closed-loop kinematic structures, focusing on three key phenomena: a) redundant boundary conditions, b) the propagation and computation of constraint forces, and c) the static equilibrium problem. In addition, aspects such as post-processing, CAD integration and user interaction were examined.

With respect to a), OpenModelica and Simscape require higher precision to achieve stable numerical solutions. In contrast, Adams and VEROSIM show superior robustness, as their numerical solvers are less sensitive to the initial position accuracy. 
Regarding b), despite employing different solvers and integrators, all four environments yield similar results when computing constraint forces in complex closed-loop systems. 
For c), OpenModelica fails to handle static equilibrium effectively, whereas Simscape, Adams, and VEROSIM demonstrate stronger capabilities in this regard.
Finally, in terms of post-processing capabilities, Adams and Simscape perform particularly well. Furthermore, Adams and also VEROSIM are characterized by good CAD integration and 3D visualization options. A unique selling point of VEROSIM is its integrated real-time capability, which allows interactive dynamic simulations. However, this requires limitations in terms of numerical quality.

In summary, it can be said that all four simulation environments examined are valuable tools with a wide range of applications. However, it should be noted that individual advantages may be achieved at the expense of other disadvantages. This should be taken into account when selecting a tool for a specific problem.


\begin{thebibliography}{6}
\bibitem {ryan}
Ryan, R. R. : ADAMS—Multibody system analysis software. In: Multibody systems handbook, 361-402 (1990).
\url{doi:10.1007/978-3-642-50995-7_21                                     }
\bibitem {poz:ach:val}
Pozzi, M., Achilli, G. M., Valigi, M. C., Malvezzi, M. : Modeling and simulation of robotic grasping in simulink through simscape multibody. Frontiers in Robotics and AI, 9, 873558 (2022).
\url{doi:10.1007/978-3-642-50995-7_21                                     }
\bibitem{rossmann2012}
Rossmann, J., Schluse, M., Schlette, C., Waspe, R. : Control by 3D Simulation – A New eRobotics Approach to Control Design in Automation. Intelligent Robotics and Applications, Berlin (2012).
\url{doi: 10.1007/978-3-642-33515-0_19                                       }
\bibitem {col:cha:van}
Collins, J., Chand, S., Vanderkop, A., Howard, D. : A review of physics simulators for robotic applications. IEEE Access, 9, 51416-51431 (2021).
\url{doi:10.1109/ACCESS.2021.3068769                                           }
\bibitem {ere:tas:tod}
Erez, T., Tassa, Y., Todorov, E. : Simulation tools for model-based robotics: Comparison of bullet, havok, mujoco, ode and physx. In: 2015 IEEE international conference on robotics and automation (ICRA) (pp. 4397-4404). IEEE (2015).
\url{doi:10.1109/ICRA.2015.7139807                                         }
\bibitem {yoo:son:lee}
Yoon, J., Son, B., Lee, D. : Comparative study of physics engines for robot simulation with mechanical interaction. Applied Sciences, 2023, 13. Jg., Nr. 2, S. 680. (2023).
\url{doi:10.3390/app13020680                                       }

\bibitem {verosim}
Osterloh, T. : Domänenübergreifende Simulation starrer Körper für Digitale Zwillinge. Dissertation (2023).
\url{https://publications.rwth-aachen.de/record/988270}
\bibitem{feat}
Featherstone, R. : Rigid body dynamics algorithms. Springer (2014).
\bibitem {bara}
Baraff, D. : Linear-time dynamics using Lagrange multipliers. In:
Proceedings of the 23rd annual conference on Computer graphics and interactive
techniques ACM, 1996, S. 137–146 (1996).
\url{doi:10.1145/237170.237226                                         }
\bibitem {ode}
Smith, R. : Open Dynamics Engine user guide (2006).

\bibitem {tod:ere:tas}
Todorov, E., Erez, T., Tassa, Y. : Mujoco: A physics engine for model-based control. In: 2012 IEEE/RSJ international conference on intelligent robots and systems (pp. 5026-5033). IEEE (2012).
\url{doi:10.1109/IROS.2012.6386109                                       }

\bibitem {mujoco}
Todorov, E. : A convex, smooth and invertible contact model for trajectory optimization. In 2011 IEEE International Conference on Robotics and Automation (pp. 1071-1076). IEEE(2011).
\url{doi:10.1109/ICRA.2011.5979814                                       }

\bibitem {GSTIFF}
Gear, C. : Simultaneous numerical solution of differential-algebraic equations. In: IEEE transactions on circuit theory, 18(1), 89-95. (1971).
\url{doi:10.1109/TCT.1971.1083221                                     }
\bibitem {adams}
Adams Solver User's Guide (2023). \url{https://documentation-be.hexagon.com/bundle/Adams_2023.3_Adams_Solver_User_Guide/raw/resource/enus/Adams_2023.3_Adams_Solver_User_Guide.pdf#M4.9.15703.hheader.INTEGRATOR}
\bibitem {simscape}
Simscape Multibody User's Guide (2024). \url{https://www.mathworks.com/help/pdf_doc/sm/sm_ug.pdf}
\bibitem {dassl}
Petzold, L. R. : Description of DASSL: a differential/algebraic system solver (No. SAND-82-8637; CONF-820810-21). Sandia National Labs., Livermore, CA (1982).
\bibitem {equil}
García de Jalón, J., Bayo, E., García de Jalón, J., Bayo, E. : Static Equilibrium Position and Inverse Dynamics. Kinematic and Dynamic Simulation of Multibody Systems: The Real-Time Challenge, 201-242 (1994).
\url{doi:10.1007/978-1-4612-2600-0                                   }
%




\end{thebibliography}
\end{document}